\documentclass[10pt,twocolumn,letterpaper]{article}

\usepackage[pagenumbers]{cvpr} 

\usepackage[dvipsnames]{xcolor}

\definecolor{cvprblue}{rgb}{0.21,0.49,0.74}
\usepackage[pagebackref,breaklinks,colorlinks,allcolors=cvprblue]{hyperref}
\usepackage{multirow}
\usepackage{graphicx}
\usepackage{gensymb}
\usepackage{amssymb}
\usepackage{pifont}
\newcommand{\cmark}{\ding{51}}%
\newcommand{\xmark}{\ding{55}}%
\usepackage{subcaption}
\usepackage[export]{adjustbox} 
\usepackage[accsupp]{axessibility}  

\def\paperID{5} 
\def\confName{USM3D2025}
\def\confYear{2025}

\newcommand*\samethanks[1][\value{footnote}]{\footnotemark[#1]}

\title{Turin3D: Evaluating Adaptation Strategies under Label Scarcity in Urban LiDAR Segmentation with Semi-Supervised Techniques
}

\author{
    Luca Barco$^{1,2}$\thanks{Equal contribution} \quad
    Giacomo Blanco$^{2}$\samethanks \quad
    Gaetano Chiriaco$^{2}$\samethanks \quad
    Alessia Intini$^{1}$ \quad \\
    Luigi La Riccia$^{1}$ \quad 
    Vittorio Scolamiero$^{3}$ \quad
    Piero Boccardo$^{1}$ \quad
    Paolo Garza$^{1}$ \quad
    Fabrizio Dominici$^{2}$
    \\\\
    $^{1}$Politecnico di Torino \quad
    $^{2}$LINKS Foundation \quad
    $^{3}$Sapienza Università di Roma
}

\begin{document}
\maketitle
\begin{abstract}
3D semantic segmentation plays a critical role in urban modelling, enabling detailed understanding and mapping of city environments. In this paper, we introduce Turin3D: a new aerial LiDAR dataset for point cloud semantic segmentation covering an area of around $1.43\:km^2$ in the city centre of Turin with almost $70M$ points. We describe the data collection process and compare Turin3D with others previously proposed in the literature. We did not fully annotate the dataset due to the complexity and time-consuming nature of the process; however, a manual annotation process was performed on the validation and test sets, to enable a reliable evaluation of the proposed techniques.
We first benchmark the performances of several point cloud semantic segmentation models, trained on the existing datasets, when tested on Turin3D, and then improve their performances by applying a semi-supervised learning technique leveraging the unlabelled training set.
The dataset will be publicly available to support research in outdoor point cloud segmentation, with particular relevance for self-supervised and semi-supervised learning approaches given the absence of ground truth annotations for the training set.
\end{abstract} 
\section{Introduction}
\label{sec:intro}
Accurate 3D semantic segmentation is a fundamental task in urban mapping, enabling applications such as infrastructure monitoring, city planning, and environmental analysis. Aerial LiDAR (Light Detection and Ranging) technology has become an essential tool for acquiring large-scale, high-resolution 3D data in urban environments, offering detailed geometric representations of buildings, roads, vegetation, and other key elements of the urban landscape. However, despite the increasing availability of aerial LiDAR data, the number of publicly available labelled datasets designed specifically for semantic segmentation remains limited. 
\begin{figure}[t]
    \centering
    \begin{subfigure}[b]{0.49\columnwidth}
        \includegraphics[width=\textwidth,valign=t]{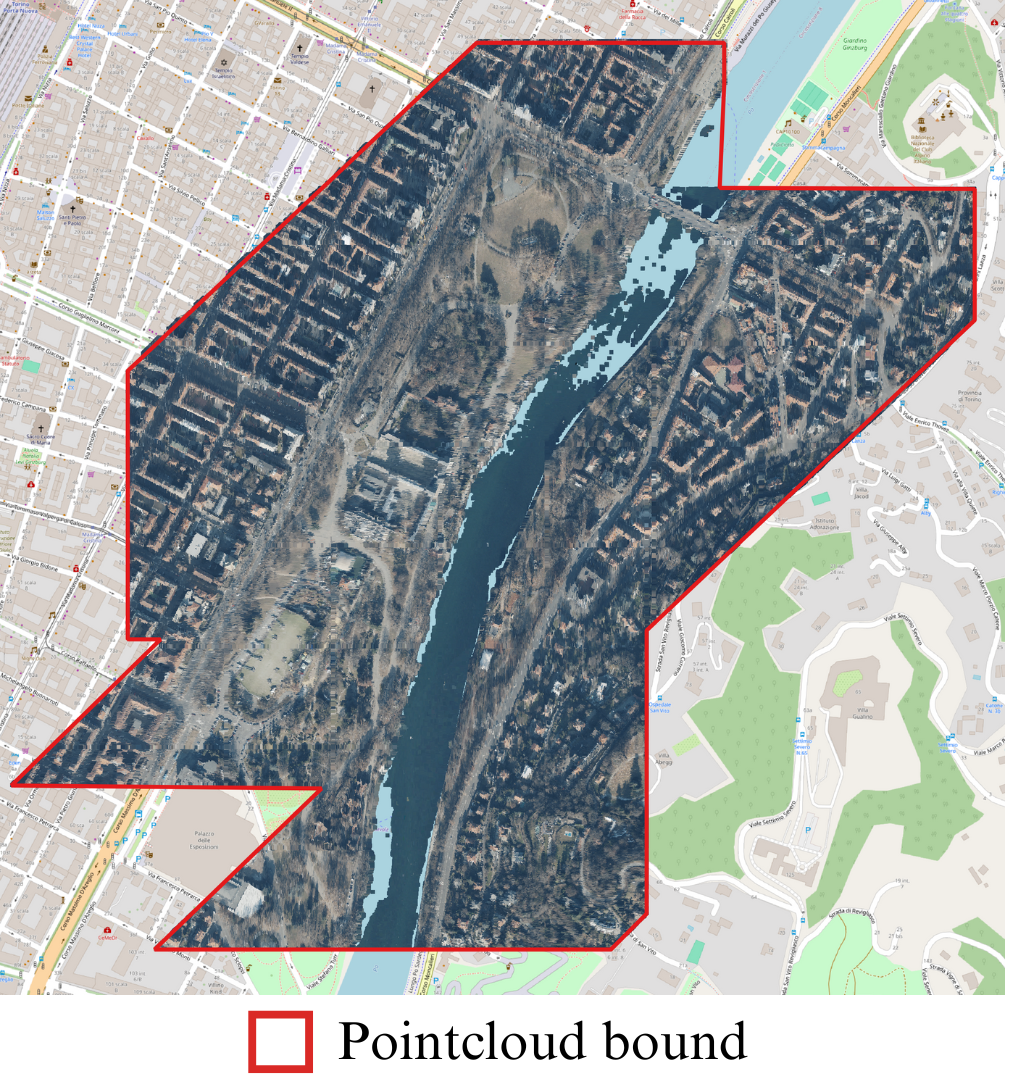} 
        \caption{Acquired RGB point cloud and its perimeter}
        \label{fig:main_pc}
    \end{subfigure}
    \hfill 
    \begin{subfigure}[b]{0.49\columnwidth}
        \includegraphics[width=\textwidth,valign=t]{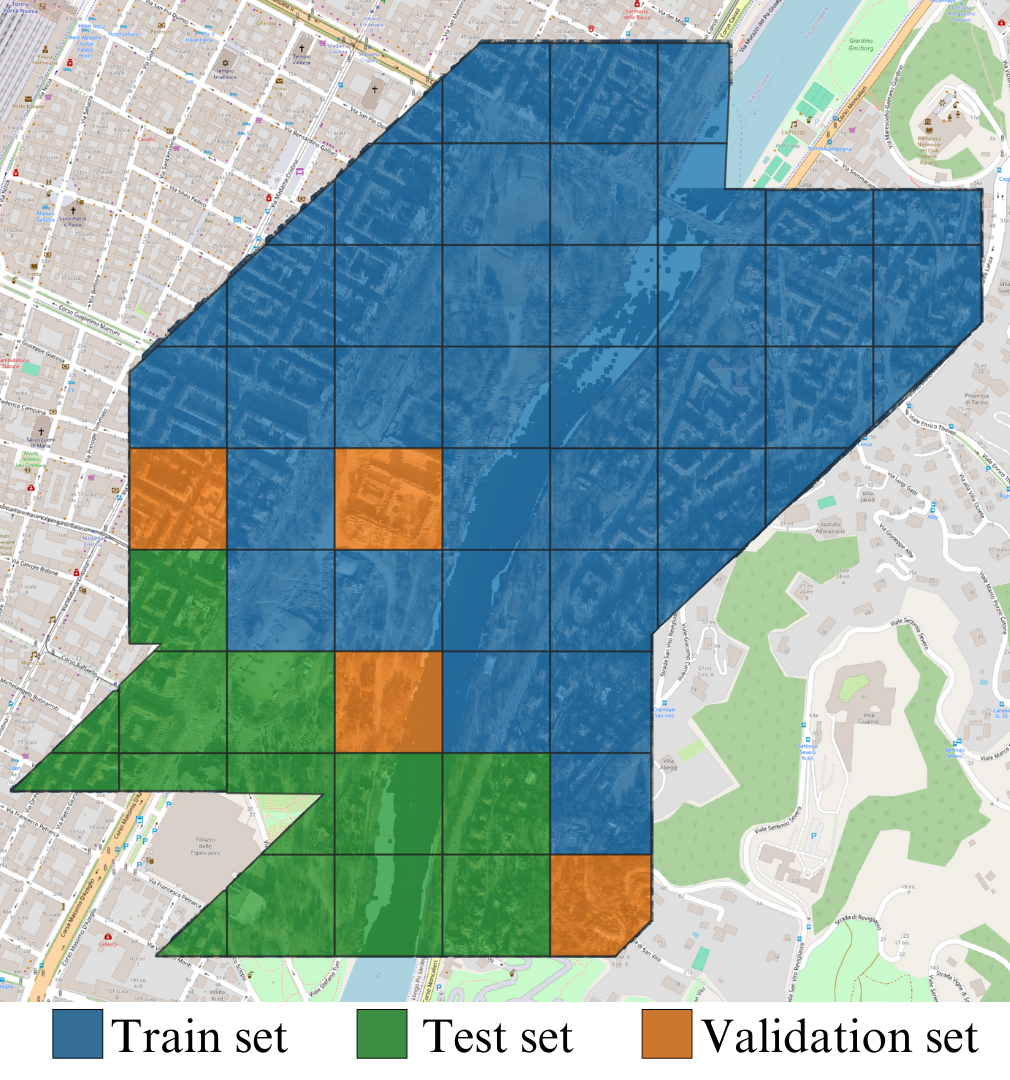} 
        \caption{Dataset split in train, validation and test set}
        \label{fig:main_split}
    \end{subfigure}
    \caption{Turin3D point cloud whole extent and subdivision in blocks}
    \label{fig:main}
\end{figure}
In this work, we introduce Turin3D: a new aerial LiDAR dataset for 3D semantic segmentation, covering an urban area of approximately $1.43\: km^2$ in the city centre of Turin, Italy. The dataset was collected using an airborne LiDAR system capable of high-density point cloud acquisition, ensuring a detailed and precise representation of complex urban structures. Unlike datasets derived from terrestrial LiDAR, which are constrained by occlusions and perspective limitations, aerial LiDAR provides a complete top-down view of the urban scene, making it particularly well-suited for applications requiring large-scale mapping and monitoring.


The dataset is divided into three subsets: training, validation, and test. While the validation and test sets have been manually annotated to provide semantic labels and quantitative performance metrics, the training set remains unlabelled, given the prohibitive cost and time-intensive nature of the process.

To assess the usability of the dataset and establish benchmark results, we conduct experiments using popular deep learning models for 3D point cloud segmentation, such as Point Transformer \cite{PointTrf} and RandLA-Net \cite{hu2020randla}. We first evaluate these models under both fully supervised conditions (leveraging existing annotated datasets) assessing their generalization capabilities on Turin3D. Then, we benchmark the best-performing architecture under semi-supervised conditions, where it is forced to learn from a training set where ground truth data is not available but instead artificially generated soft labels are used. 

The main contributions of this work can be summarised as follows.
\textit{(i)} Introduction of a new publicly available aerial LiDAR dataset for 3D semantic segmentation in urban environments, with high-resolution point clouds collected over a dense city centre \footnote{\url{https://huggingface.co/datasets/links-ads/Turin3D}}. 
\textit{(ii)} Benchmarking of popular segmentation models, evaluating their performance under different supervision settings, including scenarios without annotated training data.

By providing a new dataset and a thorough evaluation of deep learning models in different supervision regimes, this work aims to support future research in urban-scale 3D point cloud segmentation and promote the development of data-efficient learning approaches capable of leveraging partially annotated datasets.

The rest of the paper is organized as follows: Section \ref{sec:related} provides a review of related works in point cloud semantic segmentation, introducing existing datasets and popular methodologies.
Section \ref{sec:dataset} provides an overview of how the dataset was collected, the taxonomy proposed and annotation process carried out. In Section \ref{sec:methodology}, the applied methodology is explained in detail, covering the experimental setup of the different supervision settings.
Section \ref{sec:results} presents the experimental results and performance evaluation of the proposed approach. Finally, Section \ref{sec:conclusions} discusses the significance of the findings and potential avenues for future research.

\section{Related Works}
\label{sec:related}

Accurate representation of the urban environment is critical for a variety of applications, including urban planning, environmental modelling, infrastructure monitoring, energy consumption estimation, and evaluation of the green energy potential. Many of these analyses cannot be effectively conducted using only two-dimensional data, such as images or cartographic information. Therefore, incorporating the vertical dimension through three-dimensional (3D) data enhances the structural and semantic characterization of urban areas. The combination of the availability of 3D data and the application of complex algorithms, i.e. Artificial Intelligence deep learning models,  enables the development of advanced applications and analyses that can support decision-making in urban development and sustainability.

\subsection{Datasets for Urban Mapping}

The development of accurate urban digital twins relies on high-quality 3D datasets for training and validation. These datasets are commonly represented as point clouds, consisting of sets of XYZ-points, typically acquired through LiDAR scanning or photogrammetric reconstruction. Alternative representations include meshes, which define surfaces through connected polygons, and volumetric models that partition space into regular grid cells (voxels). The majority of publicly available datasets in this domain provides point clouds.

The scanning modality is a key factor that affects the information encoded by the data \cite{LIDARvs3D, UAVfor3D}. Ground-based (terrestrial) LiDAR provides highly detailed scans of building facades and street-level infrastructure but offers limited coverage of roof structures. Mobile LiDAR systems, mounted on vehicles, efficiently capture both road infrastructure and building facades along vehicle-accessible paths. Unmanned Aerial Vehicle (UAV) based systems enable flexible data collection from various heights and angles, particularly useful for capturing building roofs and areas inaccessible to ground vehicles. Aerial methods, conducted from aircraft, cover the largest areas most efficiently but provide lower detail of vertical surfaces. Photogrammetric reconstruction offers a cost-effective alternative to LiDAR by generating 3D point clouds from overlapping photographs, though potentially with lower geometric accuracy in some situations.

The datasets present in literature evidence considerable diversity in their acquisition methodologies, optical perspectives and colors, and class taxonomies.

\paragraph{UAV-Based Datasets}
SensatUrban \cite{Hu_2021_CVPR} is an urban-scale photogrammetric point cloud dataset containing almost three billion points annotated with a taxonomy of $13$ classes. It covers approximately $7.6\:km^2$ of urban landscape in three UK cities. The data were collected through aerial surveys, with automated flight paths pre-planned in a grid pattern.
The collected data were then processed using commercial software, which applies Structure from Motion (SfM) and dense image matching techniques for point cloud reconstruction. \\
Similarly, Hessigheim 3D \cite{Hessigheim3D} is a dataset used for semantic segmentation of 3D point clouds and textured meshes, data were acquired from a LiDAR system and cameras integrated on the same Unmanned Aerial Vehicle (UAV) platform. The data were collected in the village of Hessigheim, Germany, and cover an area of about $0.19\:km^2$ with over $125M$ points.
A distinctive feature of this dataset is its high spatial resolution; in fact, the point cloud features a density of about $800$ $points/m^2$. The entire point cloud has been manually labeled following a taxonomy of $11$ semantic classes.

\paragraph{Aerial Datasets}
Several datasets employ aerial acquisition methods. SUM \cite{sum2021} covers an area of $4\:km^2$ in Helsinki using airplane-mounted cameras, providing both meshes and point clouds with $6$ semantic classes. The mesh was generated from oblique aerial images with a GSD (Ground Sampling Distance) of $7.5\:cm$, acquired in 2017 using a multi-camera system mounted on an aircraft, while reconstruction was performed with techniques of aerial triangulation, dense image matching and surface reconstruction.

The FRACTAL (FRench ALS Clouds from TArgeted Landscapes)\cite{FRACTAL} dataset is a large-scale LiDAR dataset designed for the semantic 3D segmentation of heterogeneous landscapes. It consists of $100,000$ point clouds acquired by Airborne LiDAR Scanning (ALS) and covers a total area of $250\:km^2$ in five regions of France. This dataset was constructed using open-source data from \textit{the Institut national de l'information géographique et forestière (IGN)}.
This dataset includes $9261$ million points with an average density of $37$ $points/m^2$, and a semantic annotation is provided for $11$ classes.

Swiss3D \cite{Swiss3D} is a large-scale dataset designed for semantic segmentation of 3D point clouds acquired by drone photogrammetry. The dataset covers a total area of $2.7\: km^2$ spread over three Swiss cities and follows a five-class taxonomy.
The data was collected by a multi-rotor drone following dual-grid flight paths, resulting in denser and more complete point clouds than those obtained with LiDAR sensors.

STPLS3D \cite{stpls3d} is a large-scale dataset designed for semantic segmentation derived from aerial photogrammetry, covering more than $7\: km^2$ of landscapes and including $18$ semantic categories. 
The novelty of this dataset compared to others is that it combines real-world data acquired from UAVs with three synthetic versions generated through a procedural pipeline. This approach addresses common challenges in real-world data collection and annotation, such as class imbalance and heterogeneous point quality. Synthetic data were obtained through a generation pipeline that mimics the real photogrammetric acquisition process by simulating UAV flights over virtual environments. These environments were built using open geospatial data and procedural modelling tools, enabling the creation of realistic 3D point clouds that remain compatible with real-world data while eliminating the need for manual annotations.

\paragraph{Mobile Datasets}
Toronto 3D \cite{toronto3d} is a large-scale urban outdoor point cloud dataset for 3D semantic segmentation of urban environments; it is acquired through a Mobile Laser Scanning system (MLS) in Toronto, Canada and it covers $1\:km$ of urban streets comprising approximately $78.3$ million points classified into $8$ categories.
\newline
\newline
While these datasets could be used for urban mapping applications, aerial LiDAR datasets collected in dense urban environments remain relatively scarce. This limited availability poses challenges for developing and benchmarking algorithms specifically tailored for large-scale urban analysis.

\begin{table*}[htbp]
    \centering
    \begin{tabular}{lccccccc}
       
        \textbf{Dataset} & \textbf{Year} & \textbf{\# points} & \textbf{Classes} & \textbf{RGB} & \textbf{Intensity} & \textbf{Area} & \textbf{Sensor}\\
        \hline
        SensatUrban \cite{Hu_2021_CVPR} & 2020 & 2847M & 13 & \cmark & \xmark & 7.64 Km\textsuperscript{2} & UAV Photogrammetry \\

        Swiss3D \cite{Swiss3D} & 2020 & 226M & 5 & \cmark & \xmark & 2.7 Km\textsuperscript{2} & UAV Photogrammetry  \\
        
        Toronto 3D \cite{toronto3d} & 2020 & 78M & 8 & \cmark & \xmark & 1 Km\textsuperscript{2} & MLS \\

        Hessigheim \cite{Hessigheim3D} & 2021 & 74M & 11 & \cmark & \cmark & 0.19 Km\textsuperscript{2} & ULS \\
      
        SUM \cite{sum2021} & 2021 & 19M & 6 & \cmark & \xmark &4 Km\textsuperscript{2} & Aerial photogrammetry \\
      
        STPLS3D (Real) \cite{stpls3d} & 2022 & 150M & 6 & \cmark & \xmark & 1.27 Km\textsuperscript{2} & Aerial Photogrammetry \\

        
        FRACTAL \cite{FRACTAL} & 2024 & 9261M & 7 & \cmark & \cmark & 250 Km\textsuperscript{2}  & ALS \\
        
        \hline
        Turin3D (Ours) & 2025 & 69M & 6 & \cmark & \cmark & 1.43 Km\textsuperscript{2} & ALS \\
        \hline
    \end{tabular}
    \caption{Comparison of Turin3D dataset with the representative datasets for 3D semantic segmentation in urban scenarios MLS: Mobile Laser Scanning system, ULS: Unmanned Laser Scanning system, ALS: Airborne Laser Scanning system.}
    \label{tab:dataset-comparison}
\end{table*}

\subsection{3D Semantic Segmentation}

Deep learning methods for semantic segmentation of 3D point clouds in urban environments aim to extract hierarchical and spatially meaningful features, assigning each point to a specific semantic category. These approaches can be categorized into four main paradigms: \textit{projection-based}, \textit{discretization-based}, \textit{point-based}, and \textit{hybrid methods} \cite{Survey3DPC}.

Projection-based and discretization-based methods transform the point cloud into a structured representation, such as a 2D image or a voxel grid, where conventional deep learning techniques can be applied. The segmentation results are then reprojected onto the original point cloud. While these approaches leverage well-established CNN architectures, they introduce discretization artifacts and may lose fine geometric details. In contrast, point-based methods work directly on raw, unordered point clouds, preserving geometric precision but facing challenges due to the irregular distribution of points, which makes the application of standard convolutional operations less straightforward.
Traditional convolutional networks struggle with sparse 3D data due to high computational costs and loss of sparsity, known as the Submanifold Dilation Problem. Submanifold Sparse Convolutional Networks (SSCN) address this by introducing Sparse Convolutions (SC) and Submanifold Sparse Convolutions (SSC) \cite{SPCONV}. SC optimizes computation by assuming zero values for non-active sites, while SSC preserves the input sparsity structure, ensuring efficient feature extraction without unnecessary expansion. This approach improves point cloud processing efficiency, maintaining spatial continuity and minimizing resource waste.
One of the first methods specifically designed for direct point cloud processing is PointNet \cite{qi2017pointnetdeeplearningpoint}, which applies shared Multi Layer Perceptrons (MLPs) to each point independently and aggregates global information through a symmetric pooling function. This architecture ensures permutation invariance and computational efficiency but has limitations in capturing fine-grained local geometric structures. To overcome this, PointNet++ \cite{qi2017pointnetdeephierarchicalfeature} introduces a hierarchical feature extraction mechanism, recursively applying PointNet to spatially partitioned subsets of points. This allows the model to capture multi-scale local features while maintaining global context.

For large-scale point clouds, where computational efficiency is a key concern, RandLA-Net \cite{hu2020randla} has been proposed. It employs a random sampling strategy to reduce point density while integrating a local feature aggregation module, attentive pooling, and dilated residual blocks to compensate for information loss due to downsampling. This approach enables efficient processing while preserving relevant spatial details, making it suitable for large-scale outdoor scenes.

Other methods integrate principles from convolutional neural networks or transformer-based architectures to enhance feature learning. KPConv (Kernel Point Convolutions) \cite{KPConv} replaces traditional MLP-based processing with learnable kernel points, allowing continuous convolutional operations that better capture local spatial relationships. However, this comes with a higher computational cost. An alternative approach is Point Transformer \cite{PointTrf} and its variants, which utilize self-attention mechanisms to model long-range dependencies. By dynamically weighting interactions between points, these architectures improve the capture of both local and global contextual information, offering a flexible framework for point cloud segmentation.

Furthermore, frameworks have been developed and shared by authors to easily train and evaluate models using different datasets. In the context of this work, Open3D \cite{Zhou2018} has been used since it provides implementations of all the above-mentioned models.
\section{Turin3D Dataset}
\label{sec:dataset}

The following section details the steps taken to create the \textit{Turin3D} dataset. It first covers data acquisition using the \textit{Leica CityMapper-2} sensor, then explains the 3D reconstruction process combining LiDAR and aerial imagery. Additionally, this section details the chosen semantic taxonomy, annotation workflow, data partitioning strategy, and key statistics, providing an overview of the dataset’s composition.

\begin{figure*}[t]
    \centering
    \includegraphics[width=.75\textwidth,valign=t]{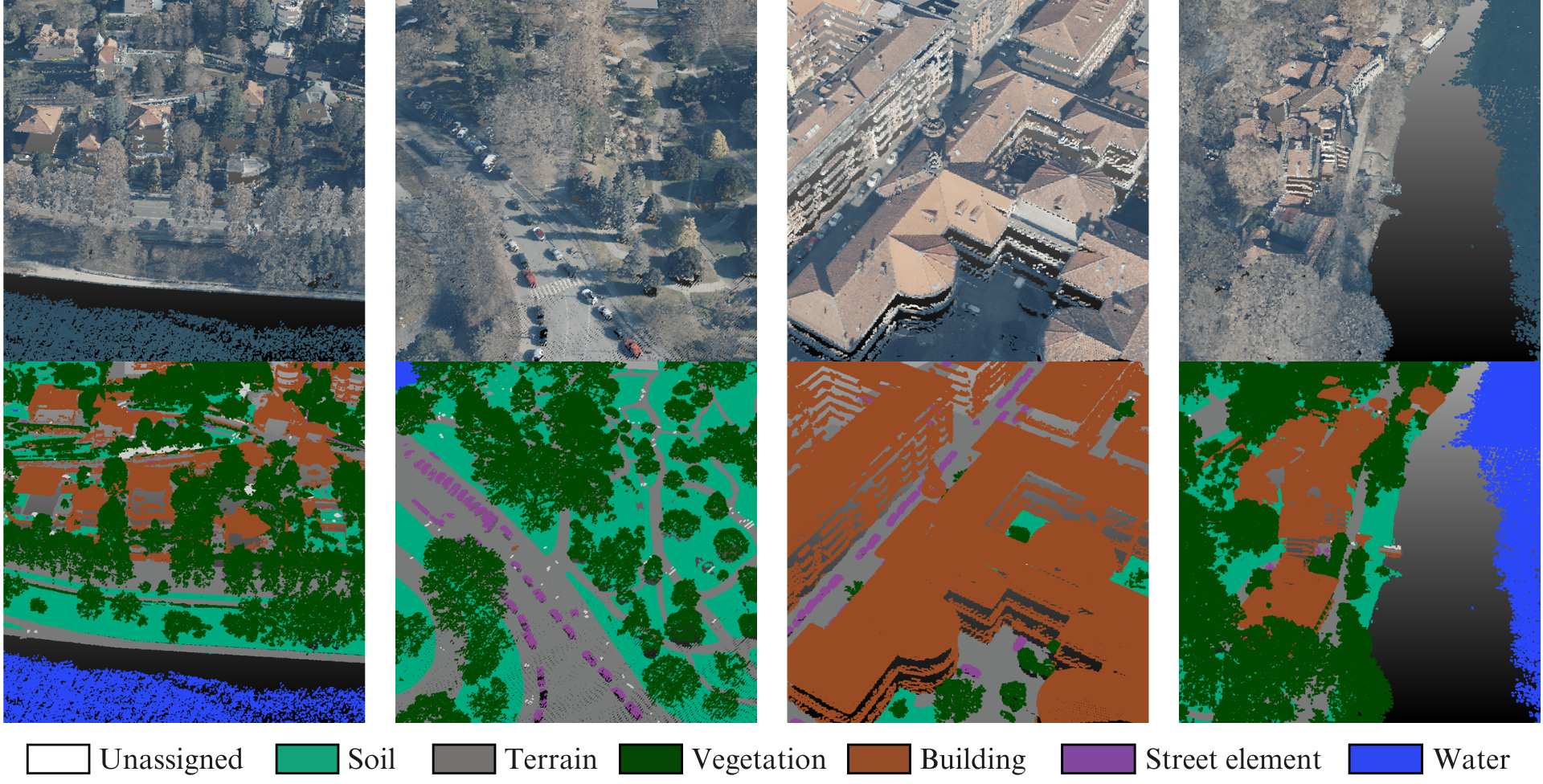} 
    \caption{Close-in views of Turin3D. Top row displays the scenes in RGB coloring, bottom row shows the same areas with points colored according to their assigned class labels.}
    \label{fig:annotation}
\end{figure*}

\subsection{LiDAR Acquisition}
\label{sec:lidar_acquisition}

Turin3D was acquired using the \textit{Leica CityMapper-2}, an airborne hybrid sensor that combines optical imagery and LiDAR scanning. The aerial survey was conducted on 28-29 January 2022, over a large area in the metropolitan city of Turin, Italy. 
The LiDAR component of the system operated with a conical scanning pattern, enabling the capture of vertical surfaces from multiple directions. The LiDAR acquisition was performed at an altitude of approximately $1\:km$, with a scanning angle of $20\degree$, resulting in a point density ranging from $30$ to $40$ $points/m^2$. This density ensures a detailed geometric reconstruction of both ground and above-ground structures, making the dataset well-suited for urban mapping applications. Simultaneously, optical imagery was acquired using a combination of nadir and oblique cameras. A total of $20,291$ images were collected, with each acquisition point capturing one nadir and four oblique images. The photogrammetric dataset features a Ground Sampling Distance (GSD) of $5\:cm$, with an $80\%$ longitudinal and $60\%$ lateral overlap, ensuring high-resolution coverage of the area. The system is equipped with two different cameras: \textit{Camera NIR Lens 71}, used for nadir and multi-spectral acquisition, and \textit{Camera RGB Lens 112/145}, used for oblique imagery.

\subsection{3D Point Cloud Processing}
\label{sec:3d_reconstruction}

The raw LiDAR data and aerial images were processed using \textit{Agisoft Metashape 2.1.0} and \textit{nFrames SURE 5.2}. These tools enabled the derivation of dense point clouds, 3D meshes, orthophotos, and Digital Terrain and Surface Models. The fusion of LiDAR and photogrammetric data aimed to add RGB features, compensate for occlusions and improve vertical surface reconstruction, particularly in high-density urban environments. A sample of the final, colourized point cloud is illustrated in Figure \ref{fig:annotation}. This integration enanches both the geometric accuracy of LiDAR and the radiometric consistency of the photogrammetric data,  making it a valuable resource for urban mapping applications and benchmark studies in 3D semantic segmentation.


\subsection{Dataset Description}
\label{sec:data_description}
The dataset was collected in the San Salvario district of Turin on 29 January 2022. The covered area shown in Figure \ref{fig:main_pc} spans approximately $1.43\:km^2$ and is made up of $69,591,759$ points. The entire area was divided into $57$ blocks, each roughly $25,000\:m^2$  in size. 
The number of points per block varies significantly depending on the location of the point cloud. This variation is due to the diverse environments within the chosen area. The west side of the area shows a more urban and residential landscape. The central part is more vegetated, featuring a park, historic buildings, and a river. This area features an overall lower point density due to the scarcity of tall buildings and the limitations of LiDAR in accurately capturing water bodies. The east side is the most diverse, featuring a hilly terrain with a mix of vegetation and large houses. the heterogeneity of the landscape enhances the dataset’s value, as it captures a wide range of urban environments despite being limited to a single city.

The point cloud data is stored in the standard LAS $1.4$ format, which provides a structured framework for encoding each point with its attributes. Each point contains XYZ coordinates, intensity values, return number, number of returns, scan direction, scan angle, GPS time, and RGB color values. 

\subsection{Semantic Labels Taxonomy}
\label{sec:turin_taxonomy}
The definition of semantic labels followed these principles: \textit{(i)} each class must be distinguishable from the others, with high heterogeneity between classes and high homogeneity inside a class, \textit{(ii)} each label class should add value for following downstream tasks and analysis, particularly for urban area planning and green applications. We decided to adopt a taxonomy composed of six distinct semantic labels. Compared to other datasets, we opted for a lower number of classes to avoid labels that are too similar and difficult for human annotators to distinguish reliably.  Additionally, we included the \textit{'Unassigned'} category for points that result from noise in the acquisition and reconstruction process, as well as masses of points that are too small to classify. All points belonging to this class were not taken into account in the experiments described in Section \ref{sec:methodology}.
The following is a list of the proposed taxonomy: \textit{Unassigned}, all unidentified points; \textit{Soil}, points that make up all kinds of natural surfaces, like meadows, soil; \textit{Terrain}, points that make up artificial grounds, such as streets, sidewalks, cemented trails; \textit{Vegetation}, all points belonging to trees, shrubs, bushes, and any other kind of low and high vegetation; \textit{Building}, all points from walls, fences, barriers, residential and historic buildings; \textit{Street elements}, cars, trucks, poles, benches; \textit{Water}, points that make up all kinds of water elements, like river, water canals and pools.


The proposed taxonomy constitutes an initial attempt to systematically differentiate and categorize the primary components typically found in urban environments.

\subsection{Annotation Process}
\label{sec:annotation}
\begin{figure}
    \centering
    \includegraphics[width=0.4\textwidth]{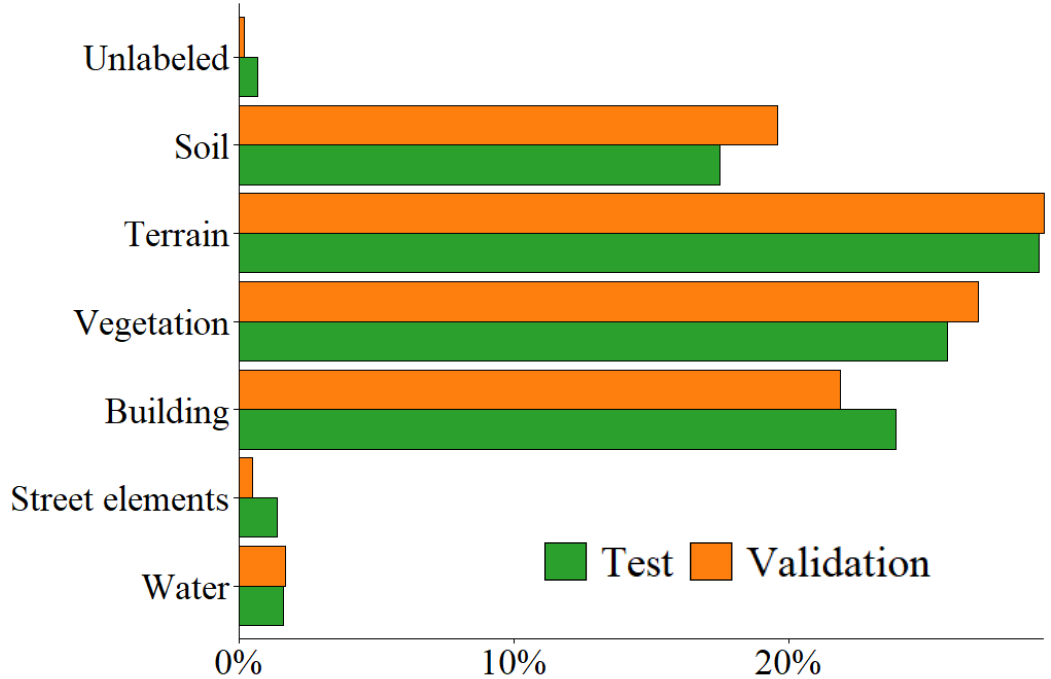} 
    \caption{Distribution of classes across test and validation sets. The percentage indicates the proportion of each of the six classes within their respective sets, not relative to the entire dataset.}
    \label{fig:class_distribution}
\end{figure}

The dataset was split into training, validation, and test sets, aiming for a point distribution as close as possible to a $70\%/10\%/20\%$ split. Only the validation and test sets were manually annotated, while the training set is used with soft labels only. This partition, shown in Figure \ref{fig:main_split}, was designed to ensure a proportional representation of the different urban settings and similar distribution of the six semantic classes, illustrated in Figure \ref{fig:class_distribution}.
The annotation process was conducted by a team of four annotators. The $17$ test and validation blocks, resulting in a total amount of almost $19M$ points, were evenly distributed among them, and each annotator worked independently in the initial phase. Each point was assigned to one of the taxonomy classes defined in Section \ref{sec:turin_taxonomy}. 

The annotation workflow began with the application of the CSF filter algorithm \cite{csf2016} to distinguish ground points from non-ground points. Ground points were then divided into natural and artificial ground, utilizing aerial imagery, RGB, and intensity features to facilitate the distinction. Non-ground points were manually classified into their respective categories, by leveraging point-wise features and available aerial images. Once all points were assigned, a final review was performed to verify coherence within local neighbourhoods and consistency with associated characteristics. To ensure consistency and reduce individual biases, a second round of review was conducted, during which the annotators collectively examined the tagged point cloud, addressing discrepancies and standardizing classification decisions across all blocks with round-table discussions.
\newline

A comparison of Turin3D with other representative datasets for 3D semantic segmentation is provided in Table \ref{tab:dataset-comparison}. With a balanced class taxonomy, manually annotated points and a diverse urban landscape acquired with high-precision aerial LiDAR sensors, Turin3D offers a valuable benchmark for evaluating point cloud segmentation models under real-world conditions

\section{Methodology}
\label{sec:methodology}

\subsection{Problem Formulation}
This research addressed the challenge of semantic segmentation of urban 3D point clouds across different urban environments, focusing on two approaches: transfer learning and semi-supervised learning with pseudo-labelling.

Let $\mathcal{D}_S = \{(x_i^S, y_i^S)\}_{i=1}^{N_S}$ represent the source domain composed of $N_S$ point clouds from literature datasets, where $x_i^S \in \mathbb{R}^{P_i \times F}$ denoted a point cloud with $P_i$ points of $F$ features, and $y_i^S \in \mathcal{C}^{P_i}$ denoted point-wise labels from class set $\mathcal{C}$ = \{\textit{Unassigned, Soil, Terrain, Vegetation, Building, Street Element, Water}\}, according to the taxonomy proposed in Section \ref{sec:turin_taxonomy}. The literature datasets included SensatUrban \cite{Hu_2021_CVPR}, DELFT SUM \cite{sum2021}, Toronto3D \cite{toronto3d}, FRACTAL \cite{FRACTAL}, STPLS3D (Real) \cite{stpls3d}, Swiss3D \cite{Swiss3D}, and Hessigheim \cite{Hessigheim3D}, with each dataset's original classes mapped to exactly one class of $\mathcal{C}$.

Let $\mathcal{D}_T = \{(x_j^T)\}_{j=1}^{N_T^{train}} \cup \{(x_k^T, y_k^T)\}_{k=1}^{N_T^{val}} \cup \{(x_l^T, y_l^T)\}_{l=1}^{N_T^{test}}$ represent the Turin3D dataset with unlabeled training and labeled validation and test sets, consisting of $N_T^{train}$, $N_T^{val}$ and $N_T^{test}$ point clouds, respectively.

Three semantic segmentation architectures $\mathcal{A}=$ \{RandLA-Net\cite{hu2020randla}, PointTransformer\cite{PointTrf}, SparseConv\cite{SPCONV}\} were evaluated using two experimental approaches as described in the following sections.

\subsection{Data Augmentation}
Throughout all experiments, consistent data augmentation was applied to improve model generalization. Geometric augmentations included linear normalization to scale coordinates, point recentering along all axes, and rotation up to 30° to handle varied terrain elevations. For color, ChromaticAutoContrast, ChromaticJitter, and HueSaturationTranslation were applied to address lighting and appearance variations across datasets. Additionally, RandomHorizontalFlip was applied to x and y axes only, preserving the natural orientation of ground surfaces in urban environments.

\subsection{Transfer Learning}
The first approach addressed the fundamental challenge of generalizing to previously unseen urban environments. Each architecture $a \in \mathcal{A}$ was trained on the literature datasets to obtain $\theta_a^* = \arg\min_{\theta} \mathcal{L}(f_{\theta}^a, \mathcal{D}_S)$

These models were then evaluated on the Turin3D test set $\mathcal{D}_T^{\text{test}}$. For these experiments, the feature set $F = \{x, y, z, R, G, B\}$ was used, excluding intensity since it was not available across all literature datasets.

The transfer learning experiments evaluated whether models trained on existing literature datasets could effectively generalize to the unseen urban environment of Turin without any domain-specific adaptation.

\subsection{Semi-Supervised Learning with Iterative Pseudo-Labeling}
The second approach leveraged the large amount of unlabeled data in Turin3D through an iterative pseudo-labeling strategy. Based on the transfer learning results, the best-performing architecture on the Turin3D validation set was selected, i.e., $a^* = \arg\max_{a \in \mathcal{A}} \text{mIoU}(f_{\theta_a^*}, \mathcal{D}_T^{\text{val}})$.

The selected model was used to generate predictions on the unlabeled Turin3D training set $\mathcal{D}_T^{train}$. Each point was assigned the class label with the highest confidence score, but only if that score exceeded a class-specific confidence threshold $\tau_c$. This filtering resulted in a set of high-confidence pseudo-labels $\hat{y}_j^T = f_{\theta_{a^*}^*}(x_j^T)$ for a subset of points in the training set.

The confidence threshold for each class was calculated as a weighted average of confidence scores from predictions on $\mathcal{D}_T^{\text{train}}$:
$$
    \tau_c = \sum_{u \in \mathcal{U}_c} u \cdot \frac{n_{u,c}}{\sum_{v \in \mathcal{U}_c} n_{v,c}}
$$

where $\tau_c$ was the confidence threshold for class $c$, $\mathcal{U}_c$ was the set of unique confidence values observed for points predicted as class $c$, $u$ was a specific unique confidence value, $n_{u,c}$ was the count of points predicted as class $c$ with confidence value $u$, and $\sum_{v \in \mathcal{U}_c} n_{v,c}$ was the total number of points predicted as class $c$.

A new instance of the selected architecture was trained from scratch on $\mathcal{D}_T^{\text{train}}$ using the pseudo-labelled points: $\theta^{**} = \arg\min_{\theta} \mathcal{L}(f_{\theta}^{a^*}, \{(x_j^T, \hat{y}_j^T)\}_{j \in \text{confident}})$.
For these trainings, an expanded feature set $F = \{x, y, z, R, G, B, Intensity\}$ was used, since intensity values were available in the Turin3D dataset.

Two iterative refinement approaches were implemented to progressively improve pseudo-label quality. The first approach, Fixed Confidence Thresholds, maintained the same class-specific confidence thresholds across all iterations. This strategy allowed assessment of whether iterative training alone could enhance performance without threshold adjustment. The second approach, Adaptive Confidence Thresholds, recalculated confidence thresholds after each iteration based on the model's evolving performance and confidence distributions. This acknowledged that as the model adapted to the target domain, the optimal confidence thresholds might shift. Thresholds were systematically adjusted based on performance metrics from the previous iteration, creating a bootstrapping mechanism that progressively refined both the model and its pseudo-labelling criteria.

\begin{table*}[t!]
\resizebox{\textwidth}{!}{%
\begin{tabular}{cccccccccclcc}
{\color[HTML]{000000} } &
  {\color[HTML]{000000} } &
  \multicolumn{8}{c}{{\color[HTML]{000000} \textbf{$\mathcal{D}_T^{test}$ (Turin3D)}}} &
   &
  \multicolumn{2}{c}{{\color[HTML]{000000} \textbf{$\mathcal{D}_S$ Test}}} \\ \cline{3-10} \cline{12-13} 
{\color[HTML]{000000} \textbf{Model}} &
  {\color[HTML]{000000} \textbf{Augmentation}} &
  {\color[HTML]{000000} \textit{Soil}} &
  {\color[HTML]{000000} \textit{Terrain}} &
  {\color[HTML]{000000} \textit{Vegetation}} &
  {\color[HTML]{000000} \textit{Building}} &
  {\color[HTML]{000000} \textit{Street Elements}} &
  {\color[HTML]{000000} \textit{Water}} &
  {\color[HTML]{000000} \textbf{mIoU}} &
  {\color[HTML]{000000} \textbf{F1}} &
   &
  {\color[HTML]{000000} \textbf{mIoU}} &
  {\color[HTML]{000000} \textbf{F1}} \\ \hline
{\color[HTML]{000000} } &
  {\color[HTML]{000000} \xmark} &
  22.44 &
  43.43 &
  53.98 &
  55.08 &
  11.04 &
  0.0 &
  {\color[HTML]{000000} 30.99} &
  {\color[HTML]{000000} 43.04} &
   &
  {\color[HTML]{000000} 31.24} &
  {\color[HTML]{000000} 43.28} \\
\multirow{-2}{*}{{\color[HTML]{000000} RandLA-Net \cite{hu2020randla}}} &
  {\color[HTML]{000000} \cmark} &
  10.05 &
  43.75 &
  81.42 &
  72.36 &
  16.70 &
  8.12 &
  {\color[HTML]{000000} \textbf{38.73}} &
  {\color[HTML]{000000} \textbf{49.42}} &
   &
  {\color[HTML]{000000} \textbf{67.39}} &
  {\color[HTML]{000000} \textbf{78.59}} \\
{\color[HTML]{000000} } &
  {\color[HTML]{000000} \xmark} &
  9.40 &
  16.19 &
  22.65 &
  10.45 &
  0.0 &
  0.0 &
  {\color[HTML]{000000} 16.97} &
  {\color[HTML]{000000} 9.86} &
   &
  {\color[HTML]{000000} 14.76} &
  {\color[HTML]{000000} 19.72} \\
\multirow{-2}{*}{{\color[HTML]{000000} Point Transformer \cite{PointTrf}}} &
  {\color[HTML]{000000} \cmark} &
  1.70 &
  2.56 &
  28.87 &
  9.59 &
  0.0 &
  0.0 &
  {\color[HTML]{000000} 7.15} &
  {\color[HTML]{000000} 11.88} &
   &
  {\color[HTML]{000000} 13.87} &
  {\color[HTML]{000000} 18.05} \\
{\color[HTML]{000000} } &
  {\color[HTML]{000000} \xmark} &
  8.84 &
  0.0 &
  29.65 &
  0.0 &
  0.0 &
  0.0 &
  {\color[HTML]{000000} 6.41} &
  {\color[HTML]{000000} 20.67} &
   &
  {\color[HTML]{000000} 12.39} &
  {\color[HTML]{000000} 30.77} \\
\multirow{-2}{*}{{\color[HTML]{000000} SparseConv \cite{SPCONV}}} &
  {\color[HTML]{000000} \cmark} &
  7.73 &
  0.0 &
  31.16 &
  0.0 &
  0.0 &
  0.0 &
  {\color[HTML]{000000} 6.48} &
  {\color[HTML]{000000} 30.93} &
   &
  {\color[HTML]{000000} 12.32} &
  {\color[HTML]{000000} 31.30} \\
\hline
\end{tabular}%
}
\caption{Results for Transfer learning experiments, with and without augmentations, evaluated on both test sets of literature selected datasets ($\mathcal{D}_S$) and labeled test set of Turin3D ($\mathcal{D}_T^{test}$), considering mIou and F1 score. For Turin3D also IoU per class is reported.}
\label{tab:exps_tl}
\end{table*}

\begin{table*}[t!]
\resizebox{\textwidth}{!}{%
\begin{tabular}{cccccccccc}
\textbf{Pseudo-Label Thresholding} &
  \textbf{Iteration} &
  \textit{\textbf{Soil}} &
  \textit{\textbf{Terrain}} &
  \textit{\textbf{Vegetation}} &
  \textit{\textbf{Building}} &
  \textit{\textbf{Street Elements}} &
  \textit{\textbf{Water}} &
  \textbf{mIoU} &
  \textbf{F1} \\ \hline
\multirow{3}{*}{\begin{tabular}[c]{@{}c@{}}Fixed Confidence\\  per iteration\end{tabular}}   & 1 & 26.17 & 50.26 & 85.38 & 73.94 & 17.77 & 17.88 & 45.23 & 57.67 \\
                                                                                             & 2 & 32.26 & 34.38 & 86.52 & 66.50 & 27.55 & 0.0 & 41.20 & 63.16 \\
                                                                                             & 3 &  26.29     &  32.64     &  68.60     &  55.74     &  7.32     &  0.0    & 31.76       & 51.49      \\
                                                                                             \hline
\multirow{3}{*}{\begin{tabular}[c]{@{}c@{}}Adaptive Confidence\\ per iteration\end{tabular}} & 1 & 26.17 & 50.26 & 85.38 & 73.94 & 17.77 & 17.88 & 45.23 & 57.67 \\
                                                                                             & 2 & 30.28 & 52.29 & 87.80 & 77.68 & 19.69 & 23.27  & \textbf{48.49} & 61.12 \\
                                                                                             & 3 & 32.89 & 50.88 & 87.62 & 69.92 & 18.01 & 30.51 & 48.30 & \textbf{74.45} \\ \hline
\end{tabular}%
}
\caption{Results for experiments with Semi-supervised learning with fixed and adaptive confidence per iteration, using RandLA-Net with Augmentations, evaluated on test set of Turin3D ($\mathcal{D}_T^{test}$ ) considering IoU per class, mIoU and F1 score.}
\label{tab:exps_confidence}
\end{table*}

Each iteration consisted of a full training cycle on $\mathcal{D}_T^{\text{train}}$ using a new instance of the selected architecture. At the end of the iteration, the best checkpoint on $\mathcal{D}_T^{\text{val}}$ was used to generate predictions and compute new thresholds to obtain pseudo-labels for the next iteration. 

For the first iteration only, thresholds $\tau_c$ were manually adjusted for \textit{Soil} and \textit{Water} classes by $\pm0.3$: reducing soil to $0.1$ (from $0.4$) to include more points, and increasing water to $0.9$ (from $0.6$) to retain only high-confidence points. These adjustments preserved under-represented classes in the pseudo-labelled data while managing class imbalance.

\section{Results}
\label{sec:results}
\subsection{Experimental Settings}
Models were trained using NVIDIA A100 GPUs with Multi-Instance GPU (MIG) partitioning, specifically utilizing 20GB and 40GB MIG slices. Training ran for 200 epochs with a batch size of 4 and a maximum of $65,536$ points per batch element, balancing accuracy and memory constraints. An initial learning rate of $0.001$ was applied for model optimization.
All the experiments were globally evaluated using mean Intersection over Union (mIoU) and F1-score. For Turin3D, IoU per class was also considered to provide more granular performance analysis.

\subsection{Transfer Learning}
Transfer learning experiments, reported in Table~\ref{tab:exps_tl},  revealed significant performance variations across architectures when generalizing to Turin3D.

RandLA-Net performed best ($38.73$ mIoU with augmentation) but still showed a substantial drop from its performance on literature datasets ($67.39$ mIoU), highlighting cross-city generalization challenges. Data augmentation improved overall performance, especially for \textit{Vegetation} and \textit{Buildings}, though \textit{Soil} classification degraded.

Other architectures struggled significantly: Point Transformer ($7.15$ mIoU) performed worse with augmentation than without, while SparseConv achieved only $6.48$ mIoU. These models particularly struggle with street elements and water classes, often failing completely. \textit{Vegetation} appears to be the most transferable class across all models, likely due to its more consistent appearance across different urban environments. Indeed, the complete failures in class transferability across different architectures, particularly evident for \textit{Water} and \textit{Street Elements}, likely stem from significant variations in how these elements appear in different urban environments, combined with potential annotation inconsistencies between datasets. For instance, \textit{Water} features in Turin3D may have distinct geometric or reflectance properties compared to those in the training datasets, rendering them unrecognizable to models without domain-specific adaptation. Nevertheless, RandLA-Net with augmentation achieves $8.12$ IoU on Water, suggesting that data augmentation can partially mitigate this situation.

These findings established RandLA-Net as the best architecture for subsequent semi-supervised learning experiments.

\subsection{Semi-Supervised Learning}
Semi-supervised learning experiment, reported in Table~\ref{tab:exps_confidence}, were conducted using RandLA-Net with augmentation, comparing fixed and adaptive confidence thresholding strategies across multiple iterations. 
The initial results of both approaches represent a $+6.50$ mIoU improvement over the transfer learning baseline. The adaptive thresholding approach showed consistent improvement, with mIoU peaking at $48.49$ in the second iteration and F1 score reaching $74.45$ in the third iteration. \textit{Water} segmentation demonstrated marked improvement with adaptive thresholding, increasing from $17.88$ to $30.51$ IoU ($+12.63$) across iterations, while \textit{Vegetation} maintained consistently high performance around $87$ IoU and \textit{Soil} steadily improved to $32.89$ IoU. In contrast, the fixed thresholding strategy exhibited progressive performance deterioration, declining to $31.76$ mIoU by the third iteration. Notably, \textit{Water} classification completely disappeared after the first iteration with fixed thresholds, highlighting a critical limitation of this approach: low-confidence classes become progressively excluded from pseudo-labels, creating a self-reinforcing cycle of degradation.\textit{Building} and \textit{Terrain} classes showed performance fluctuations with both approaches, indicating challenges in generating consistent pseudo-labels for these classes with complex and diverse urban appearances. Results on \textit{Water} highlight the efficacy of the adaptive approach: the adaptive method successfully maintained and improved water classification, unlike the fixed approach where this class disappeared entirely.

In conclusion, the semi-supervised learning methodology with adaptive confidence thresholding yielded a $9.76$ absolute mIoU improvement over the transfer learning baseline. This demonstrates the effectiveness of leveraging unlabeled target domain data through iterative pseudo-labeling for cross-city point cloud segmentation.

\section{Conclusions and Future Works}
\label{sec:conclusions}

In this work, we introduced \textit{Turin3D}, a new aerial LiDAR dataset for urban semantic segmentation, and evaluated different learning strategies to address the challenge of label scarcity. Our experiments compared transfer learning and semi-supervised learning techniques, demonstrating how the latter methods yielded superior segmentation performance by effectively leveraging the unlabeled training set. These results highlight the potential of data-efficient learning strategies in large-scale urban point cloud analysis, where full annotation is often impractical.  

Despite these promising results, several aspects remain open for future research. First, an extended annotation effort on the training set would allow for a fully supervised benchmark, providing a more precise evaluation of different learning strategies. Additionally, future work could explore the application of more recent deep learning architectures for point cloud segmentation, potentially improving performance over the baseline models used in this study. Another important direction is the investigation of domain adaptation techniques specifically designed for point cloud segmentation, which could enhance model generalization to unseen urban environments, further reducing the reliance on extensive manual labelling. Lastly, a further step for future research could consist of expanding the proposed taxonomy to incorporate a more detailed classification of urban elements, thereby improving its descriptive power and applicability across a broader range of urban scenarios. 
By making Turin3D publicly available, we aim to support further research in semi-supervised and transfer learning strategies for urban LiDAR segmentation, providing a challenging yet realistic benchmark for the community.

\section*{Acknowledgements}
This work was carried out in the context of Horizon Europe project UP2030 (G.A. n.101096405), Space IT Up project funded by the Italian Space Agency (ASI) and the Ministry of University and Research (MUR) under contract n. 2024-5-E.0 - CUP n. C53C24000530005 and funded by the European Union - NextGenerationEU, Mission 4 Component 2 - ECS00000036 - CUP B13D21011790006
{
    \small
    \bibliographystyle{ieeenat_fullname}
    \bibliography{main}
}


\end{document}